\documentclass[a4paper,conference]{IEEEtran}

\usepackage{times}
\usepackage{epsfig}
\usepackage{graphicx}
\usepackage{amsmath}
\usepackage{amssymb}
\usepackage{tabularx}
\usepackage{svg}
\usepackage[hidelinks]{hyperref}
\usepackage[caption=false,font=footnotesize]{subfig}

\usepackage{pifont}
\usepackage{multirow, booktabs}

\newcommand{\xmark}{\ding{55}}
\newcommand{\cmark}{\ding{51}}

\begin{document}

\title{Points to Patches: Enabling the Use of Self-Attention for 3D Shape Recognition}

\author{\IEEEauthorblockN{Axel Berg\textsuperscript{1,2}, Magnus Oskarsson\textsuperscript{2}, Mark O'Connor\textsuperscript{1}}
\IEEEauthorblockA{\textsuperscript{1}Arm ML Research Lab \\
Email: \{axel.berg,mark.oconnor\}@arm.com \\
\textsuperscript{2}Centre for Mathematical Sciences, Lund University, Sweden\\
Email: \{magnus.oskarsson\}@math.lth.se}}

\maketitle

\begin{abstract}
While the Transformer architecture has become ubiquitous in the machine learning field, its adaptation to 3D shape recognition is non-trivial. Due to its quadratic computational complexity, the self-attention operator quickly becomes inefficient as the set of input points grows larger. Furthermore, we find that the attention mechanism struggles to find useful connections between individual points on a global scale. In order to alleviate these problems, we propose a two-stage Point Transformer-in-Transformer (Point-TnT) approach which combines local and global attention mechanisms, enabling both individual points and patches of points to attend to each other effectively. Experiments on shape classification show that such an approach provides more useful features for downstream tasks than the baseline Transformer, while also being more computationally efficient. In addition, we also extend our method to feature matching for scene reconstruction, showing that it can be used in conjunction with existing scene reconstruction pipelines.
\end{abstract}

\section{Introduction}
Due to the unordered nature of 3D point clouds, applying neural networks for shape recognition requires the use of permutation-equivariant architectures. The Transformer, first introduced by Vaswani et al.\ \cite{vaswani2017attention} for the task of natural language processing, is an example of such an architecture and given its recent success in the fields of image classification \cite{dosovitskiy2021image, touvron2021training}, object detection \cite{carion2020end}, video analysis \cite{neimark2021video}, speech recognition \cite{gulati2020conformer, chen2021developing, liu2021tera, berg21_interspeech}, and more, its application to 3D point clouds is a natural step of exploration. While some attempts to adopt the Transformer architecture for this task have been made \cite{lee2019set, zhao2021point, guo2021pct}, they all suffer from different weaknesses, such as a reduced receptive field and high computational cost. Inspired by the Transformer-in-Transformer architecture for image processing \cite{han2021transformer}, we propose a method that addresses both of these problems by using a two-stage attention mechanism which is able to learn more descriptive features while also lowering the number of required computations.

In summary, our main contributions are

\begin{enumerate}
\item We propose a two-stage Transformer architecture that combines attention mechanisms on a local and global scale. By sampling a sparse set of anchor points we create patches of local features. Self-attention can then be applied both on points within the patches and on the patches themselves.
\item Our experiments show that this approach gives significant uplifts compared to applying self-attention on the entire set of points, while also reducing the computational complexity.
\item We show that our proposed architecture achieves competitive results on 3D shape classification benchmarks, while requiring less floating point operations (FLOPs) compared to other methods, and that it can be used for improving 3D feature matching.
\end{enumerate}

\begin{figure}
\center
\includegraphics[width=\linewidth]{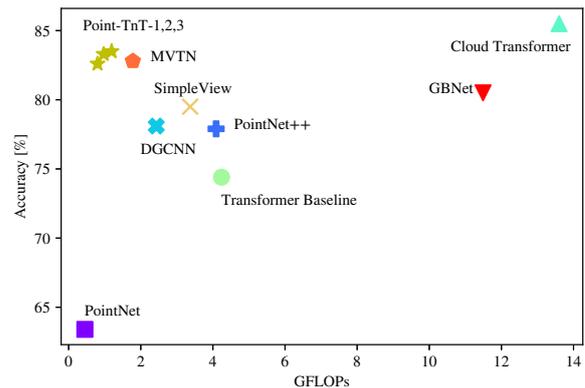}
\caption{Classification accuracy and GFLOPs on the ScanObjectNN dataset \cite{uy2019revisiting}. Our proposed Point-TnT networks provides the best trade-off between accuracy and fast inference when compared to previous methods, and significantly improves performance compared to a Transformer baseline.}
\label{pareto}
\end{figure}

\section{Related Work}

\begin{figure*}
\centering
\includegraphics[trim={0 5cm 0 4cm},clip, width=\linewidth]{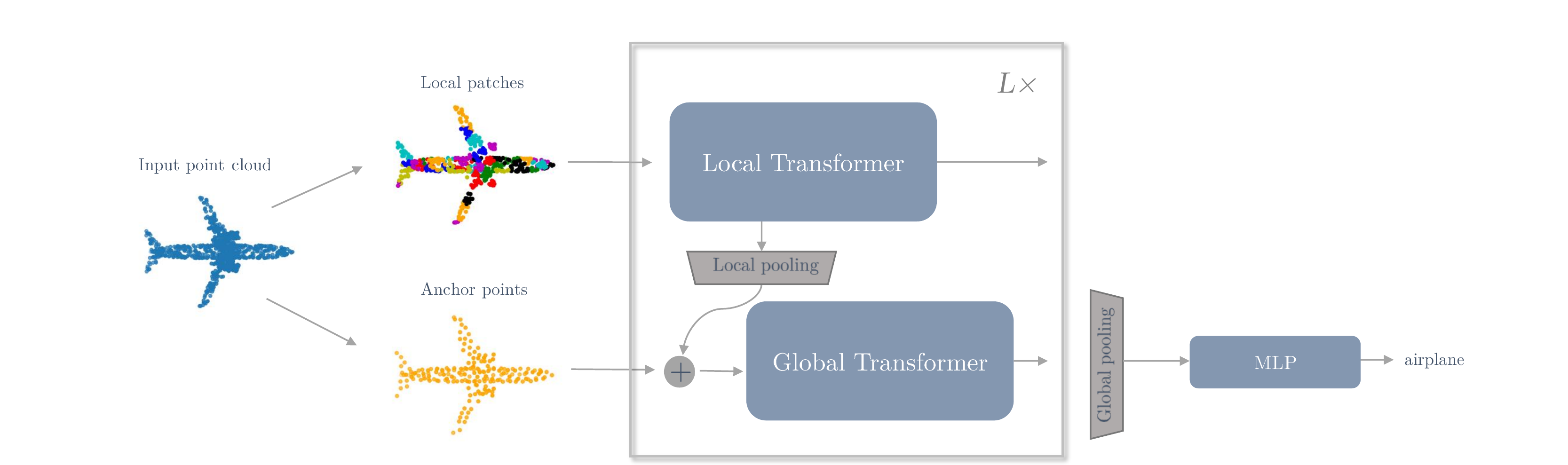}
\caption{Overview of our proposed Point Transformer-in-Transformer method. Given an input point cloud, we sample a sparse set of anchor points and then create local patches around each anchor using the $k$-nearest neighbour graph. The anchors and patches are then fed into a sequential network of local and global Transformers, each with its own self-attention mechanism. At the final layer, global and local features can be extracted and used for downstream tasks, such as 3D shape classification.}
\label{overview}
\end{figure*}

\noindent \textbf{Shape Recognition} Extracting useful features from 3D point cloud shapes requires special considerations. In contrast to images, where the pixels are naturally ordered on a grid structure, a point cloud is typically represented as an unordered set of data points, each containing three-dimensional coordinates and possibly other feature channels, such as normal vectors and color information. This prevents the direct use of classical deep learning techniques like multi-layer perceptrons (MLPs) and convolutional neural networks (CNNs) when combining features from different points, since their predictions depend on the ordering of the input features. In recent years, several paradigms have been developed to work around this problem and they can roughly be divided into three categories: volumetric, multi-view and set based approaches.

The volumetric approach, as proposed in \cite{wu20153d, maturana2015voxnet}, aims to map the points onto a three-dimensional occupancy grid of voxels, which allows for the use of CNNs for feature extraction. However, this approach suffers from poor scaling behaviour with the grid-resolution, making it intractable for processing of large point cloud scenes. This problem can be alleviated to some extent by using sparse convolutions \cite{choy20194d}, but will inevitably suffer from loss of geometric detail when points are mapped to an occupancy grid. In contrast, the multi-view approach instead maps the point clouds onto multiple two-dimensional grids, each capturing a different view of the scene. This method, first proposed in \cite{su2015multi}, has later been refined to use adaptive views that are learned at training time \cite{hamdi2021mvtn}. Others have proposed combining the multi-view approach with graph convolutional networks (GCN) in order to aggregate views over nearby view positions \cite{wei2020view}.

A family of neural networks termed Deep Sets, which allows for feature extraction on unordered sets without mapping them to any grid-like structure, was introduced by Zaheer et al.\ \cite{zaheer2017deep}. Such networks can be realized by weight sharing across all input points and the use of a symmetric aggregation function, e.g.\ the max or mean values of the sets, which enables extraction of permutation-equivariant point features. Concurrent work by Qi et al.\ \cite{qi2017pointnet} introduced PointNet, where this architecture was combined with a Spatial Transformer network \cite{jaderberg2015spatial} (not to be confused with the attention-based Transformer), that can learn to align point clouds in a common frame of reference \cite{qi2017pointnet}. This approach has later been enhanced by exploiting geometric properties of the point cloud, such as the nearest neighbour graph \cite{qi2017pointnet++, wang2019dynamic}, which enables aggregation not only over the entire point cloud, but also over local neighbourhoods of points. \\~\

\noindent \textbf{Self-Attention and the Transformer} The Transformer architecture \cite{vaswani2017attention} belongs to the family of permutation-equivariant neural networks and is therefore a natural extension of the Deep Sets architecture. By allowing all elements of the input set to attend to each other, the Transformer is able to learn interactions between elements, which greatly enhances the network ability to learn complex interactions. It was first applied to point clouds by Lee et al.\ \cite{lee2019set}, who noted that this quickly becomes infeasible as the computational requirement grows quadratically with number of input points. They addressed this problem by introducing the concept of induced set attention by allowing the points to attend to a set of learnable parameters instead of themselves.

Zhao et al.\ \cite{zhao2021point} proposed a modified Transformer architecture, which only computes local attention between a point and its nearest neighbours. However, this approach removes the main benefit of self-attention, namely that it enables a global receptive field in early layers of the network. In contrast, Guo et al.\ \cite{guo2021pct} introduced offset attention with a neighbour embedding module in order to perform attention between groups of points, where each group is represented by a local feature. This method omits local attention entirely and for segmentation tasks it also suffers from quadratic complexity in the number of input points. Finally, a third approach has been explored in \cite{chen2020dapnet}, where self-attention is computed both between points and channels within points, but only at a late stage in the network and only for the particular task of areal LiDAR image segmentation. 

Recently, Transformers have also shown great success in the domain of computer vision in the form of the Vision Transformer \cite{dosovitskiy2021image, touvron2021training}, where an image is regarded as a set of local patches, which can then be processed using a Transformer with minimal modifications. Han et al.\ \cite{han2021transformer} proposed the Transformer-in-Transformer architecture, which extends the baseline vision Transformer with pixel-wise attention within patches. Our approach is inspired by this work, in the sense that we use two branches of Transformers, one for local and one for global attention. This enables a global receptive field early in the network, while reducing the computational requirement of the self-attention operation compared to a baseline Transformer implementation.

\section{Method} \label{method}

\noindent \textbf{Preliminaries} Let $X \in \mathbb{R}^{N \times d}$ be a matrix representation of a set, containing $N$ features in $d$-dimensional space. Furthermore, following the definitions introduced in \cite{vaswani2017attention}, let $Q = X_lW_Q$, $K = X_lW_K$ and $V = X_lW_V$ denote the queries, keys and values respectively, where $W_Q, W_K, W_V \in \mathbb{R}^{d \times d_h}$ are learnable parameters and $d_h$ is the attention-head dimension. We then define the self-attention (SA) operator as

\begin{equation}
\text{SA}(X) = \text{Softmax}\Big(\frac{QK^T}{\sqrt{d_h}}\Big)V,
\end{equation}
where the softmax function is applied on each row individually. We note that the SA operation is permutation-equivariant, since for any permutation $\pi$ over the rows of $X$, we have that $\text{SA}(\pi X) = \pi \text{SA}(X)$. This property is especially useful if $X$ represents a collection of unordered points, which is common in many 3D learning scenarios.

By performing multiple SA operations, where each operator has its own learnable set of weights, in parallel and concatenating the results column-wise, we can define the multi-headed SA operator (MSA) as

\begin{equation}
\text{MSA}(X) = [\text{SA}_1(X); \text{SA}_2(X); ...; \text{SA}_h(X)] W_P,
\end{equation}
where $;$ denotes column-wise concatenation, $W_P \in \mathbb{R}^{h d_h \times d}$ is another learnable parameter and $h$ is the number of attention heads. Using the above notation, we can define the Transformer layer $T_\theta$ as

\begin{align}
\tilde{X} &= \text{MSA}(\text{LN}(X)) + X,\\
T_\theta(X) &= \text{MLP}(\text{LN}(\tilde{X})) + \tilde{X},
\end{align}
where MLP denotes a multi-layer perceptron with a single hidden layer and GELU activation  \cite{hendrycks2016gaussian}, LN is the LayerNorm \cite{ba2016layer} operation and $\theta$ is the collection of parameters for the particular layer. Here we use the pre-norm \cite{he2016deep} variant of the Transformer, where LayerNorm is applied before the MSA and MLP operations.

In order to aggregate features in a permutation-invariant fashion, we follow \cite{wang2019dynamic} and compute the maximum and average over all features and concatenate the results column-wise. This can be compactly expressed as 

\begin{equation}
\alpha (X) = [\max_i X_i; \frac{1}{N}\sum_i X_i], \quad i = 1,..., N,
\end{equation}
where $X_i$ are the rows of $X$. \\~\

\begin{figure}
  \centering
  \smallskip
  \includesvg[width=\linewidth]{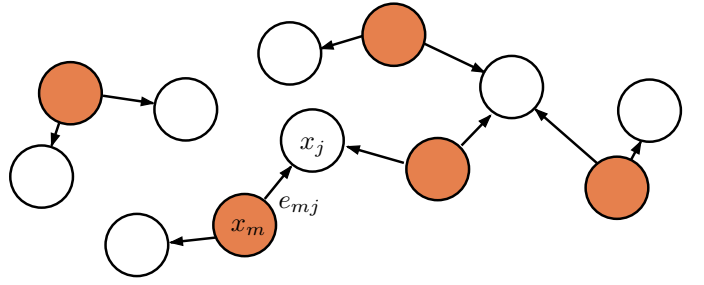}
  \caption{An example of the sparse subgraph $\mathcal{H}$, using $M = 5$ anchor points (shown in orange) and $k=2$ nearest neighbours.}
  \label{graph}
\end{figure}

\noindent \textbf{Point Transformer-in-Transformer} Now we are ready to define our main architecture. Let $\mathcal{X} =\{x_i\}_{i=1}^N$  be a set of $N$ points in three-dimensional space. A naive application of the Transformer architecture would be to apply self-attention between all points, such that each point in the set can attend to every other point. Here we instead investigate how the attention mechanism can be applied within and between local patches of points, effectively splitting the self-attention operator into two different branches. An overview of our method can be seen in Figure \ref{overview}.

First, we compute the $k$-nearest neighbour ($k$-NN) graph $\mathcal{G} = (\mathcal{V}, \mathcal{E})$, with vertices $\mathcal{V} = \{i\}_{i=1}^N$ and directed edges $\mathcal{E} \subset \mathcal{V} \times \mathcal{V}$ from vertex $i$ to $j$ if $x_j$ is one of $x_i$'s $k$-nearest neighbours. In order to create patches, we then sample a sparse set of $M$ anchor points and aggregate features from their neighbours. We do this by considering the subgraph $\mathcal{H} = (\mathcal{V}, \mathcal{E}')$ of $\mathcal{G}$, which has edges from $i$ to $j$ if $i \in \mathcal{V}' \land (i,j) \in \mathcal{E}$, where $\mathcal{V}' \subset \mathcal{V}$ and $|\mathcal{V}'| = M < N$. In practice, we create $\mathcal{H}$ by sampling a subset of points using farthest point sampling \cite{eldar1997farthest}, as shown in Figure \ref{graph}, which guarantees that the vertices in $\mathcal{V}'$ are evenly distributed across the whole point cloud. From now on, we will refer to the set $\mathcal{Y} = \{x_m\}_{m \in \mathcal{V}' }$ as the anchor points.

Since local geometry is best represented in a local frame of reference \cite{wang2019dynamic}, we then extract edge features for each anchor point as $E^m = [e_{m1}, ..., e_{mk}]^T$, where

\begin{equation}
e_{mj} = x_j - x_m, \quad (m,j) \in \mathcal{E}'.
\end{equation}

Without loss of generality, we can stack the anchor points into a matrix $Y = [x_{1}, ..., x_{M}]^T \in \mathbb{R}^{M\times 3}$, where the order of the anchor points is not important. These are then, together with their corresponding edge features, projected to higher dimensions as $Y_0 = YW_Y$ and $E^m_0 = E^mW_E$, where $W_Y \in \mathbb{R}^{3 \times d_Y}$ and $W_E \in \mathbb{R}^{3 \times d_E}$ respectively. 

The anchor and edge features are then processed using two branches of $L$ sequential Transformer blocks. The local branch computes self-attention between edge features, which enables interaction between edges within the neighbourhoods of the anchor points:

\begin{equation}
E^m_l = T^{\text{local}}_{\theta_l}(E^m_{l-1}), \quad l = 1, .., L.
\end{equation}

After each local Transformer layer, the neighbourhood features are aggregated and concatenated into a new matrix as $ E_l = [\alpha(E^1_l), ..., \alpha(E^M_l) ] \in \mathbb{R}^{M\times 2d_E} $, which is then added to each anchor point using another linear projection:

\begin{equation}
\tilde{Y}_{l-1} = Y_{l-1} + E_l W_{l}, \quad l = 1, ..., L,
\end{equation}
where $W_{l} \in \mathbb{R}^{2d_E \times d_Y}$. Now, $Y_l$ contains feature descriptors for the local patches around each anchor point. The global branch then computes attention between the patches, which enables interaction on a global scale:

\begin{equation}
Y_l = T^{\text{global}}_{\theta_l}(\tilde{Y}_{l-1}), \quad l = 1, ..., L.
\end{equation}

In order to exploit intermediate feature representations, the anchor point features from each layer are concatenated, combined using a single-layer MLP and aggregated in order to form a single global feature for the entire point cloud:

\begin{equation}
Z = \alpha(\text{MLP}([Y_1; ...; Y_L])),
\label{signature}
\end{equation}
such that $Z \in \mathbb{R}^{2d_f}$, where $d_f$ is the embedding dimension of the global feature. The global feature is then used for downstream tasks, using e.g.\ another MLP for classification. Alluding to the Transformer-in-Transformer for images \cite{han2021transformer}, we refer to our proposed method as Point Transformer-in-Transformer (Point-TnT). \\~\

\noindent \textbf{Computational Analysis} Whereas a naive transformer implementation requires $\mathcal{O}(N^2)$ computations for the SA operator, the complexity is reduced significantly by splitting the attention into local and global branches. For our method, the complexity of the local and global transformers are $\mathcal{O}(Mk^2)$ and $\mathcal{O}(M^2)$ respectively, resulting in a reduced number of self-attention operations compared to a naive Transformer implementation as long as $Mk^2 + M^2 < N^2$, a condition which can easily be satisfied by limiting the number of neighbours and anchor points.

\section{Experiments}

\subsection{Shape Classification}

\begin{table}[t]
\caption{Classification results on ScanObjectNN \cite{uy2019revisiting}. Results obtained using Protocol 2 are marked by $^*$.}
\small
\resizebox{\linewidth}{!}{
\begin{tabular}{lccc}
\toprule
\textbf{Method} & input & overall acc. & mean acc. \\
\midrule
PointNet \cite{qi2017pointnet} & 1024p & 68.2 & 63.4 \\
SpiderCNN \cite{xu2018spidercnn} & 1024p & 73.7 & 69.8 \\
PointNet++ \cite{qi2017pointnet} & 1024p & 77.9 & 75.4 \\
DGCNN \cite{wang2019dynamic} & 1024p & 78.1 & 73.6 \\
PointCNN \cite{li2018pointcnn} & 1024p & 78.5 & 75.1 \\
BGA-PointNet++ \cite{uy2019revisiting} & mask, 1024p & 80.2 & 77.5 \\
BGA-DGCNN \cite{uy2019revisiting} & mask, 1024p, & 79.7 & 75.7 \\
Cloud Transformer \cite{mazur2021cloud} & mask, 2048p & $85.5^*$ & $83.1^*$ \\
SimpleView \cite{goyal2021revisiting} & 6 views & 79.5 & - \\
GBNet \cite{qiu2021geometric} & 1024p & $80.5^*$ & $77.8^*$ \\
MVTN \cite{hamdi2021mvtn} & 12 views & $82.8^*$ & - \\
\midrule
Baseline (ours) & 1024p & \begin{tabular}{l}$74.4 \pm 1.3$  \\$75.4\pm 0.4^*$\end{tabular}  & \begin{tabular}{l}  $69.6 \pm 1.5$ \\ $71.3 \pm 1.1^*$ \end{tabular} \\
\midrule
Point-TnT (ours) & 1024p & \begin{tabular}{l}$83.5 \pm 0.1$  \\$84.6 \pm 0.5^*$\end{tabular} & \begin{tabular}{l}$81.0 \pm 1.3$  \\$82.6 \pm 1.2^*$\end{tabular}  \\
\midrule
Point-TnT (ours) & 2048p & \begin{tabular}{l}$84.2 \pm 0.9$  \\$85.0 \pm 0.9^*$\end{tabular} & \begin{tabular}{l}$81.8 \pm 0.9$  \\$83.0 \pm 0.8^*$\end{tabular}  \\
\bottomrule
\end{tabular}
}
\label{scanobjectnn}
\end{table}

We train and evaluate our model on the ScanObjectNN dataset \cite{uy2019revisiting}, which contains 14,510 real-world 3D objects in 15 categories, obtained from scans of indoor environments. This dataset is particularly challenging, since the point clouds contain cluttered backgrounds and partial occlusions. We use the hardest version of the dataset (PB\_T50\_RS), which has been altered using random perturbations, and the official 80/20 train/test split. This dataset also comes with per-point labels that can be used to segment the point cloud into the object and background categories. While some methods use the segmentation masks during training in order to learn to discard points belonging to the background, we do not exploit this information, since it requires additional computational overhead.

In the default setting, we use $M = 192$ anchor points and $k = 20$ neighbours. The embedding dimensions are chosen as $d_Y = 192$ and $d_E = 32$ for the anchor points and edges respectively, and we use a global feature embedding dimension of $d_f = 1024$ and a total of $L = 4$ sequential Transformer blocks. Following \cite{touvron2021training}, we let $d_Y / h = 64$ and consequently use $h=3$ attention heads for both Transformer branches. As in \cite{wang2019dynamic}, the final MLP used for classification contains two hidden layers of size 512 and 256 respectively, with batch normalization and dropout applied in each layer. We train our network using the standard cross-entropy loss function for 500 epochs, with a batch size of 32 using the AdamW optimizer \cite{loshchilov2018decoupled} with a weight decay of 0.1 and a cosine learning rate schedule starting at 0.001. For data augmentation, we use RSMix \cite{lee2021regularization} in addition to random anisotropic scaling and shifting. For each experiment, we train our model three times and report a 95 \% confidence interval for the mean accuracy\footnote{Code is available at \url{https://github.com/axeber01/point-tnt}}. 

We compare our method to a naive approach that applies self-attention directly on all points, which corresponds to using all points as anchors and no neighbours, thereby discarding the local branch of the network, and refer to this method simply as Baseline. When comparing to previously published methods, we use two different evaluation protocols. Protocol 1 uses the model at the last training epoch for evaluation on the test set, whereas Protocol 2 evaluates the model on the test set each training epoch and reports the best obtained test accuracy. Although Protocol 2 clearly exploits information from the test set, we present results using both protocols for better side-by-side comparisons with other methods. For further discussion of different evaluation protocols, we refer to \cite{goyal2021revisiting}.

\begin{figure*}
\centering
\subfloat{\includegraphics[width=0.25\linewidth, trim={2cm 2cm 2cm 2cm}]{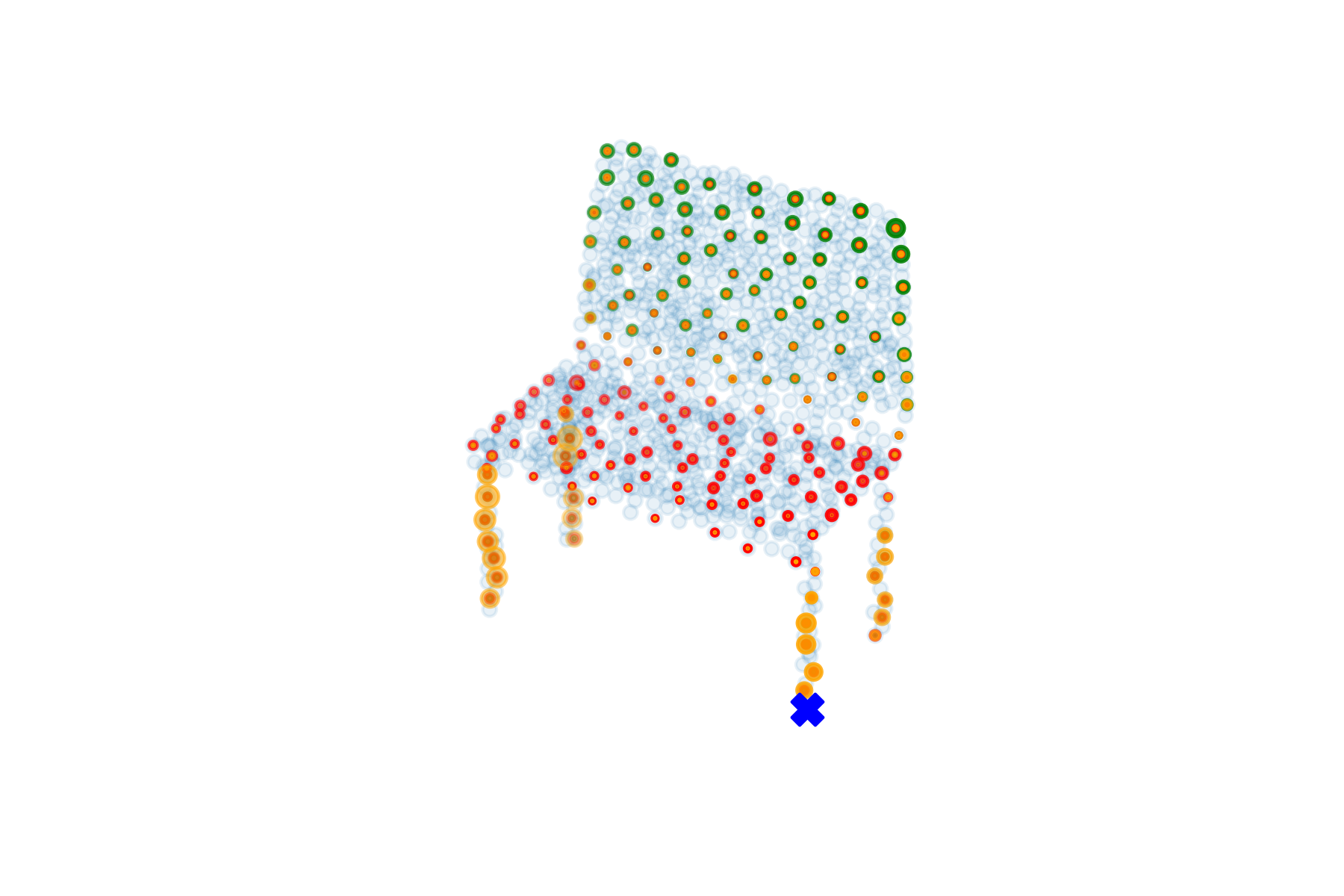}}
\subfloat{\includegraphics[width=0.25\linewidth, trim={2cm 2cm 2cm 2cm}]{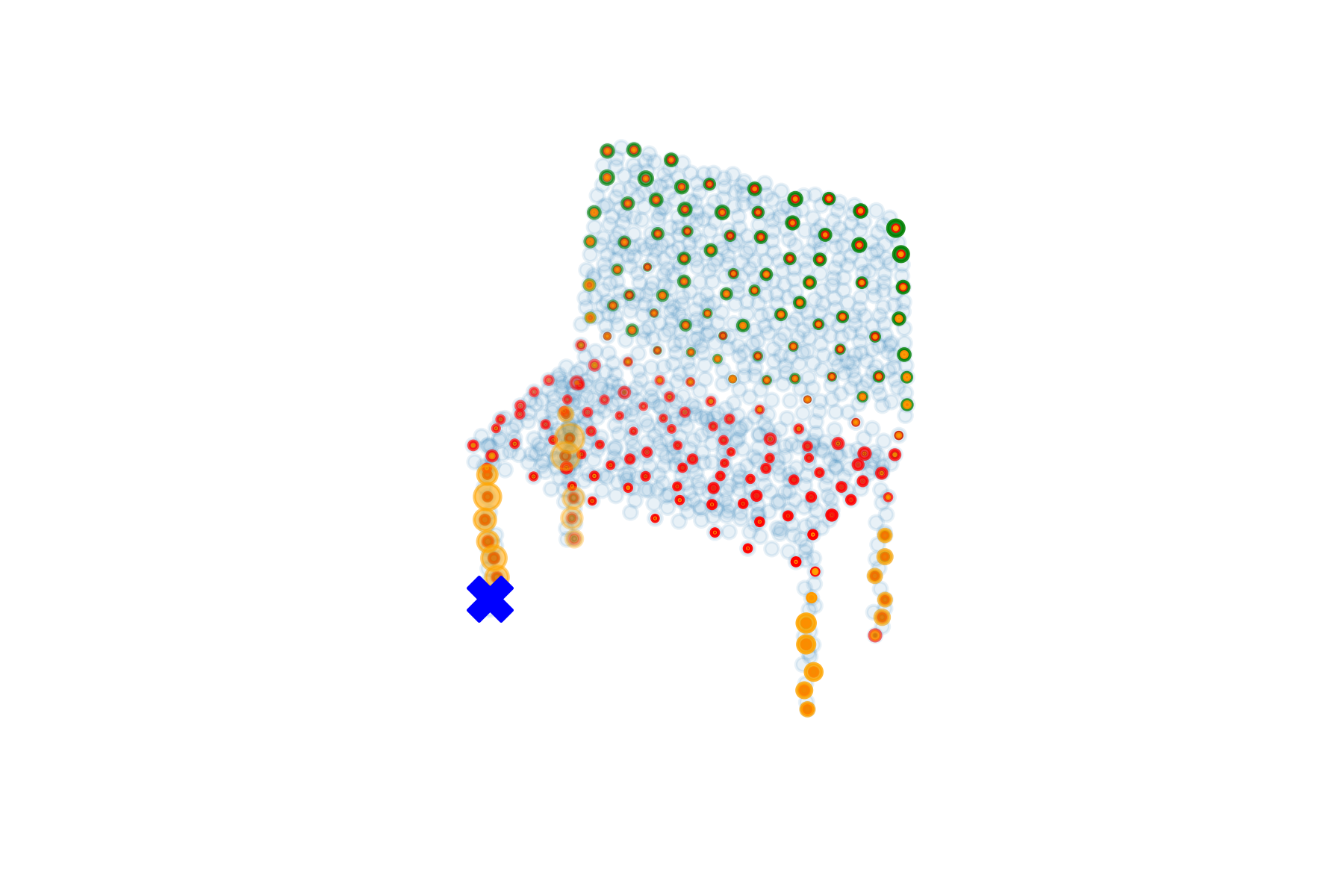}}
\subfloat{\includegraphics[width=0.25\linewidth, trim={2cm 2cm 2cm 2cm}]{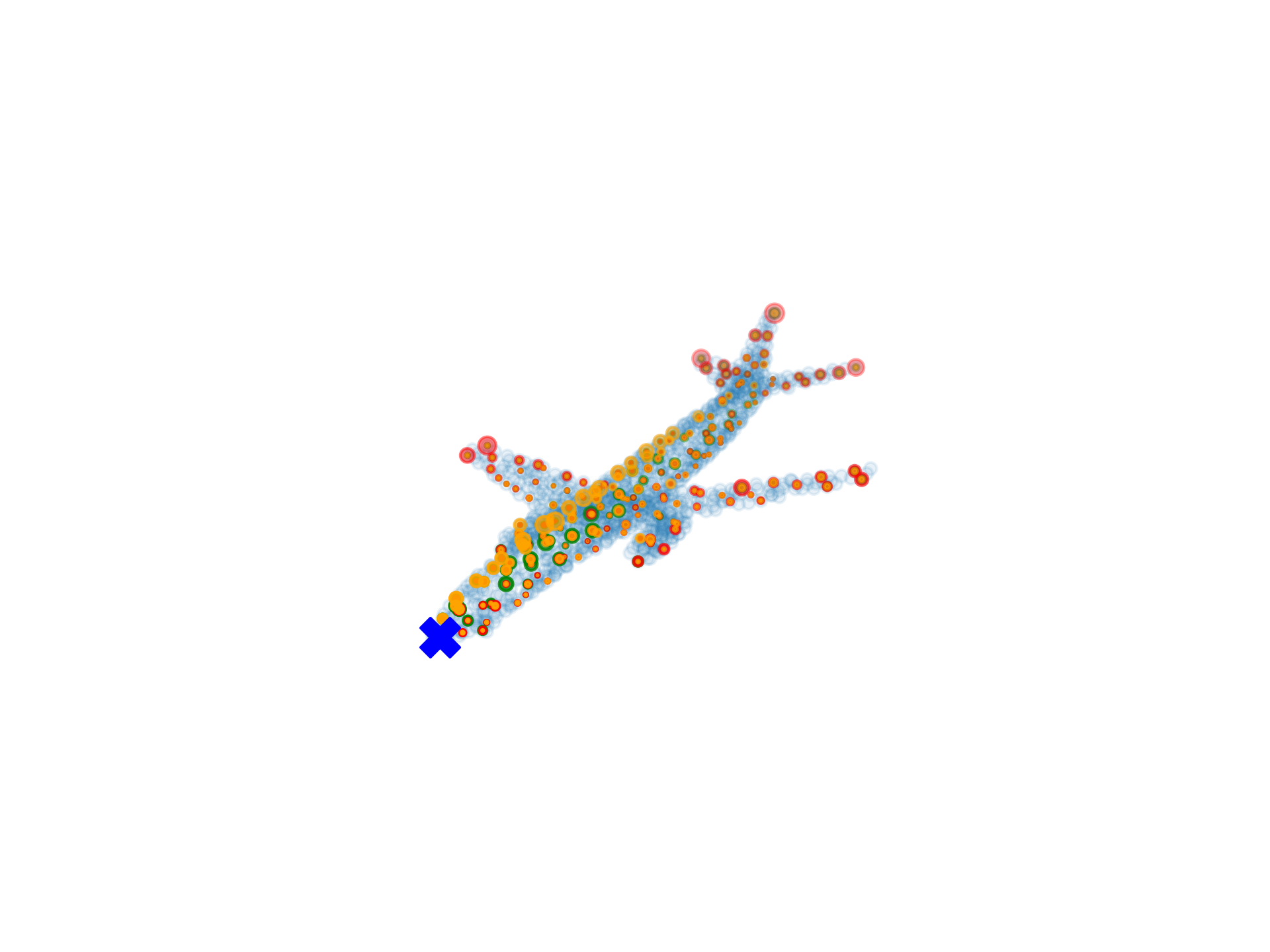}}
\subfloat{\includegraphics[width=0.25\linewidth, trim={2cm 2cm 2cm 2cm}]{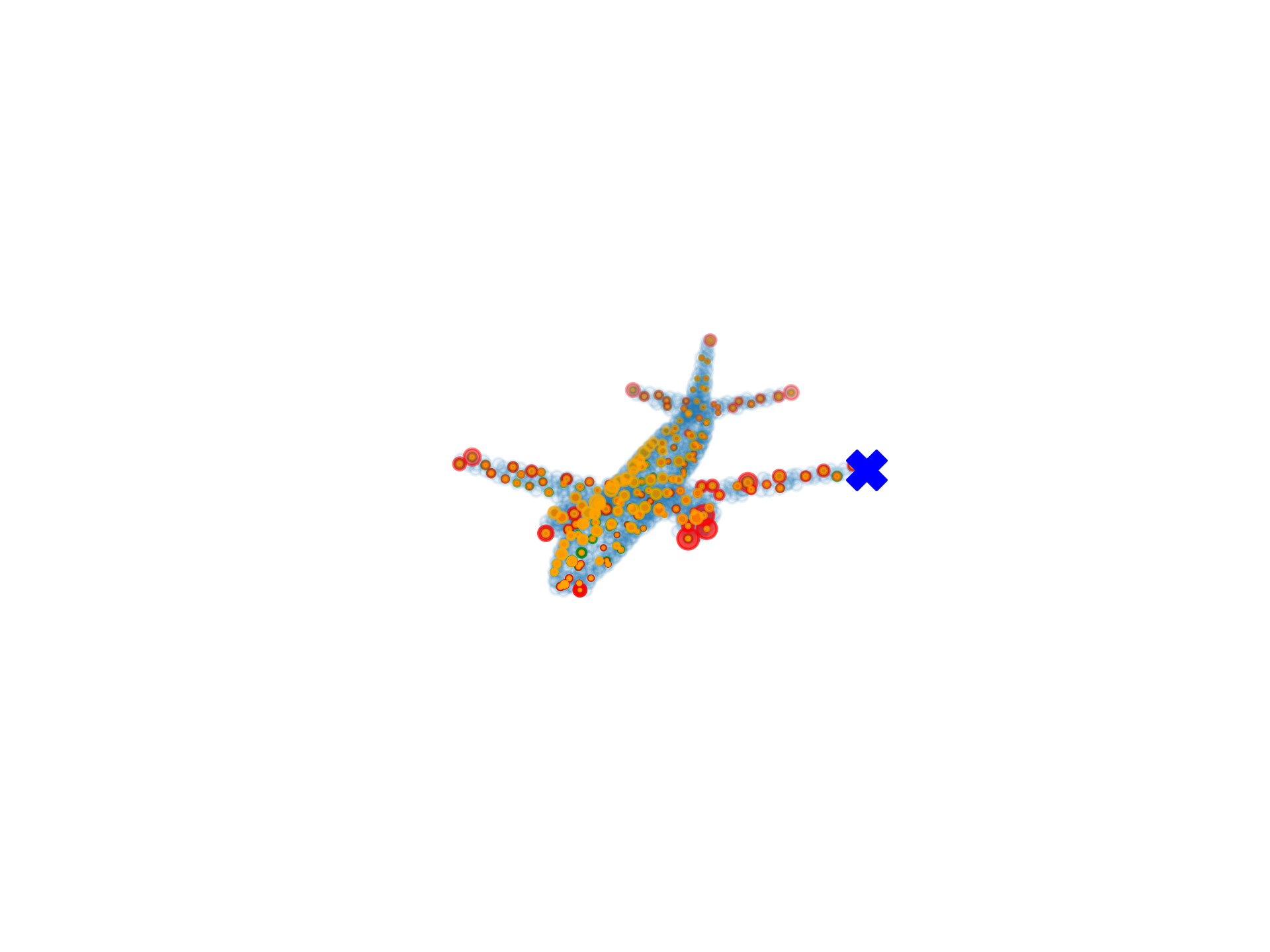}}
\hfill \\
\centering
\subfloat{\includegraphics[width=0.25\linewidth, trim={2cm 2cm 2cm 2cm}]{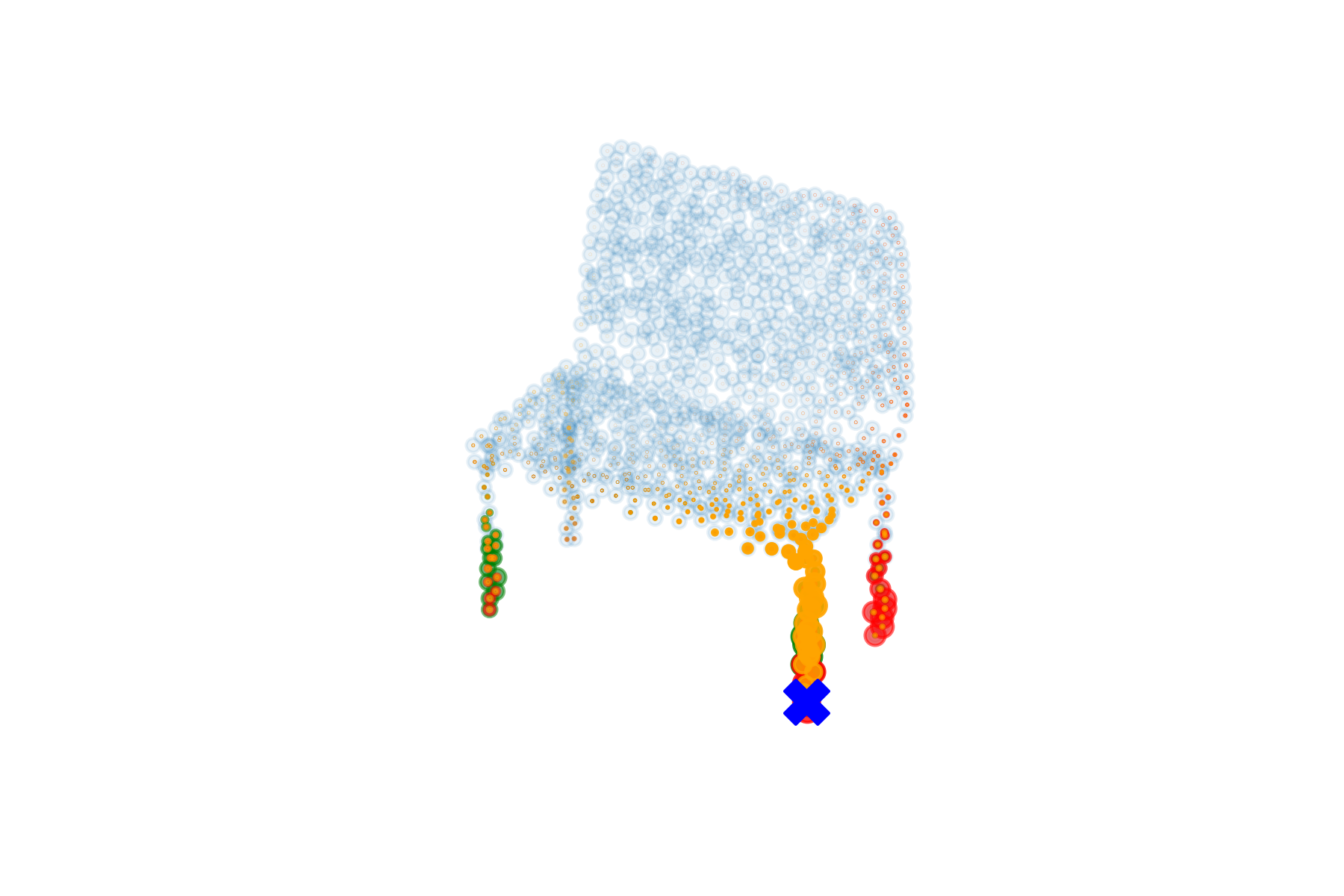}}
\subfloat{\includegraphics[width=0.25\linewidth, trim={2cm 2cm 2cm 2cm}]{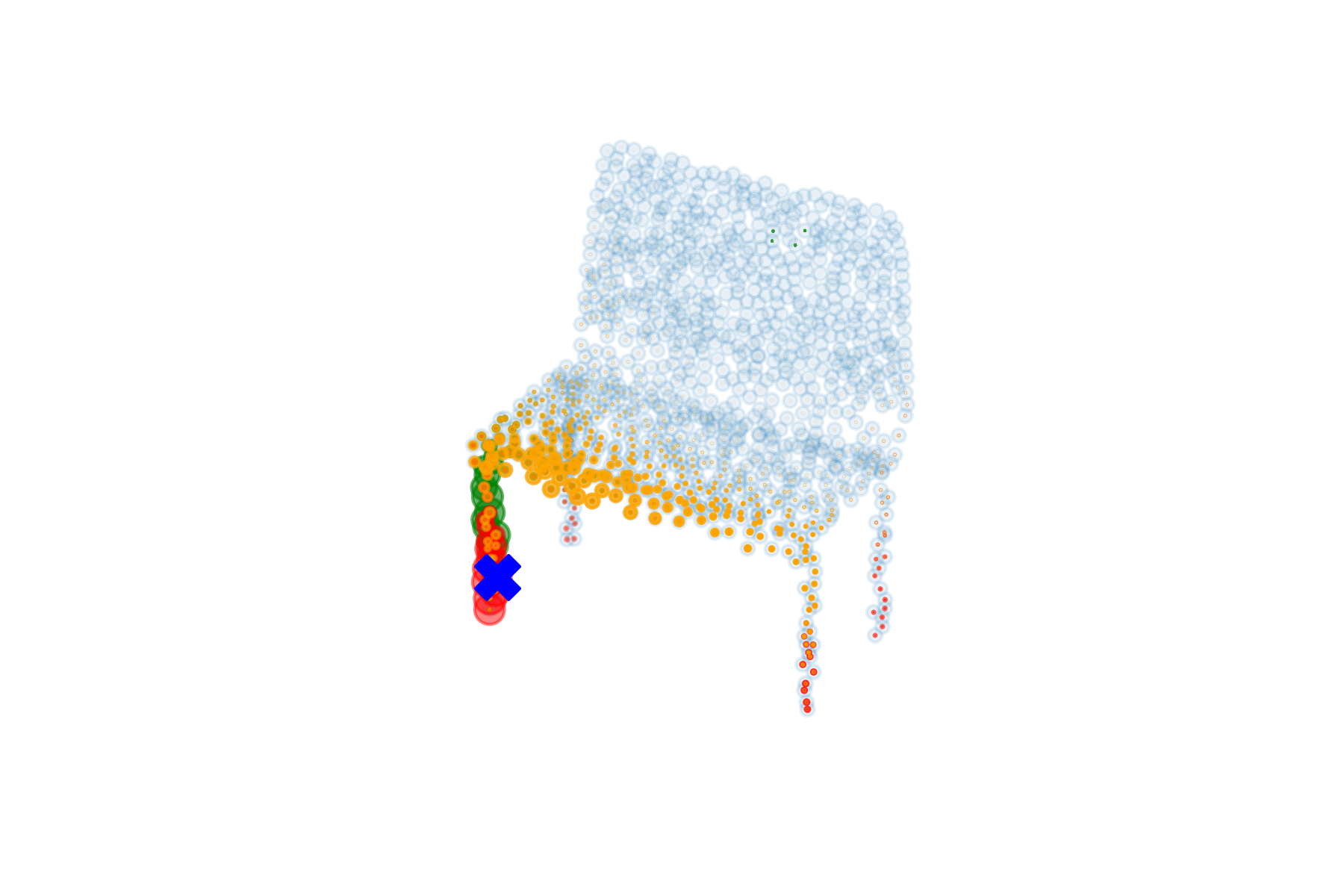}}
\subfloat{\includegraphics[width=0.25\linewidth, trim={2cm 2cm 2cm 2cm}]{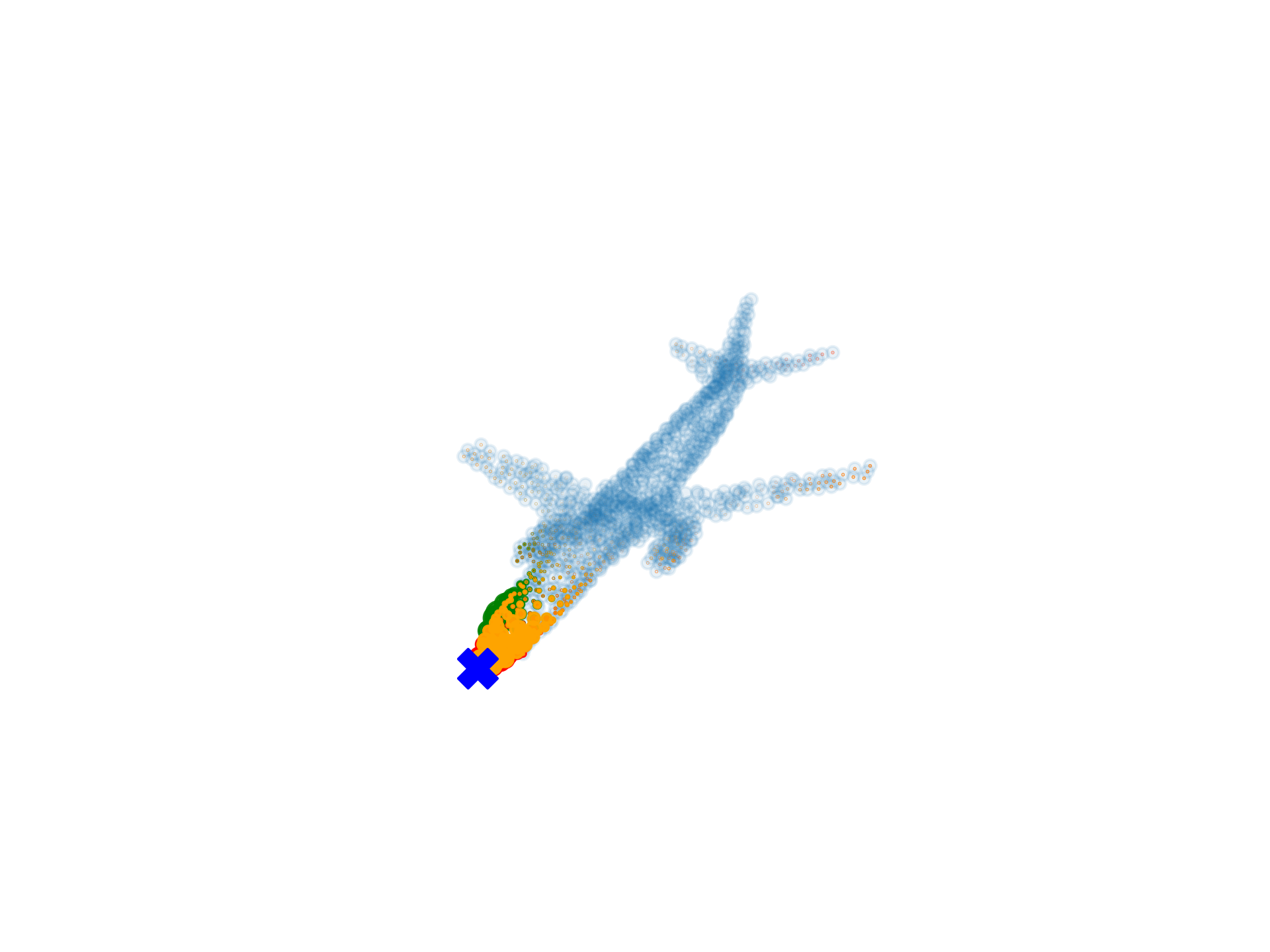}}
\subfloat{\includegraphics[width=0.25\linewidth, trim={2cm 2cm 2cm 2cm}]{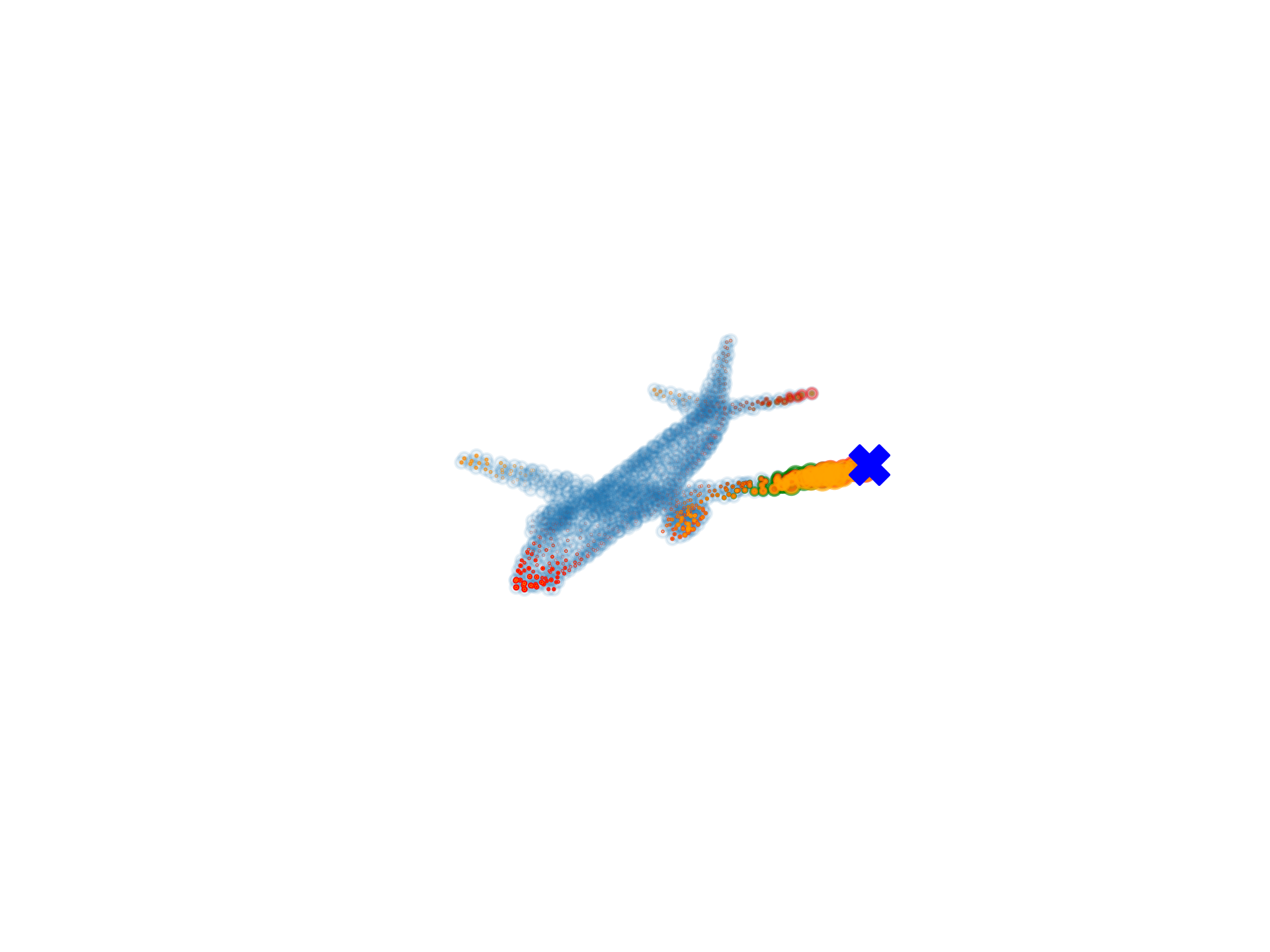}}
\hfill
\caption{Visualization of the learned attention patterns in the final Transformer layer. The blue cross indicates the point for which the attention map is computed and the red, green and yellow dots highlight the attention for the three different heads, where the size of the points are scaled proportionally to the magnitude of the attention weights. Top: our proposed method using a sparse set of anchor points. Bottom: baseline method, using all points as anchors.}
\label{attention_fig}
\end{figure*}

We compare the overall accuracy and average per-class accuracy to previously published methods in Table \ref{scanobjectnn}. Our method achieves an accuracy on par with state-of-the-art results, and the best result among the methods that do not exploit the background segmentation masks during training, outperforming all multi-view methods. Furthermore, it can be observed that using a sparse set of anchor points in conjunction with aggregation over neighbouring points performs significantly better than the baseline approach, which is not competitive on this task. This shows that global point-wise attention is not enough to capture semantic properties of objects, and that it is necessary to consider local geometric properties in order to efficiently utilize the attention mechanism.

For completeness, we also evaluate our method on the ModelNet40 dataset \cite{wu20153d}, which contains 9,843 synthetic shapes for training and 2,468 for testing, in 40 different categories. We obtain $92.6 \pm 0.2$ \% and $93.2 \pm 0.2$ \% accuracy using Protocol 1 and 2 respectively, which is competitive with other methods. 

We visualize the learned attention patterns from the different attention heads, as shown in Figure \ref{attention_fig}, using our proposed method and the naive baseline. It can be seen that when using a sparse set of anchor points, using their neighbours to form patches, the attention mechanism effectively learns to segment the point cloud into semantically meaningful parts. This allows the network to connect different parts of the point cloud when computing the global feature. However, when using the baseline approach, the attention map is not semantically meaningful. This illustrates the importance of neighbourhood context when computing self-attention, which is not used in the naive baseline approach.

\begin{table}[t]
	\caption{Model size, computational complexity and accuracy on ScanObjectNN for the baseline and our proposed method with different number of attention heads.}
	\centering
	\begin{tabular}{l|ccc}
	\toprule
	\textbf{Method} & \#params & GFLOPs & acc. \\
	\midrule
	Baseline & 3.8M & 4.33 & $74.4 \pm 1.3$ \\
	Point-TnT-1 & 1.7M & 0.80 & $82.6 \pm 0.4$ \\
	Point-TnT-2 & 2.6M & 0.97 & $83.3 \pm 0.1$ \\
	Point-TnT-3 & 3.9M & 1.19 & $83.5 \pm 0.1$ \\
	\bottomrule

	\end{tabular}
	\label{tab:dim_ablation}
\end{table}

\begin{table}[t]
	\caption{Accuracy on ScanObjectNN with and without different attention mechanisms for the baseline and our proposed method.}
	\centering
	\begin{tabular}{l|ccc}
	\toprule
	\textbf{Method }& local att. & global att. & acc. \\
	\midrule
	\multirow{2}{*}{Baseline} & & \xmark & $67.1 \pm 0.7$\\
	& & \cmark & $74.4 \pm 1.3$ \\
	\midrule
	\multirow{4}{*}{Point-TnT} & \xmark & \xmark & $79.8 \pm 0.8$ \\
	& \cmark & \xmark & $80.4 \pm 1.8$ \\
	& \xmark & \cmark & $82.9 \pm 0.1$ \\
	& \cmark & \cmark & $83.5 \pm 0.1$ \\
	\bottomrule
	\end{tabular}
	\label{tab:att_ablation}
\end{table}

\subsection{Ablation Study}

\begin{table*}[t]
	\caption{Accuracy on ScanObjectNN for different number of anchor points $M$ and nearest neighbours $k$.}
	\centering
	\begin{tabular}{lccc|ccc}
	& \multicolumn{3}{|c|}{anchors} & \multicolumn{3}{c}{neighbours} \\
	\midrule
	\multicolumn{1}{c|}{$M/k$} & 12 & 48 & 192 & 5 & 10 & 20 \\
	\multicolumn{1}{c|}{acc.} & $74.2 \pm 0.7$ & $82.5 \pm 0.8$ & $83.5 \pm 0.1$ & $79.7 \pm 1.6$ & $83.5 \pm 1.5$ & $83.5 \pm 0.1$ \\
	\bottomrule
	\end{tabular}
	\label{tab:anchor_ablation}
\end{table*}

\noindent \textbf{Model scaling} In order to highlight the trade-off between model size, computational complexity and classification accuracy, we ablate the number of attention heads (and scale the embedding dimension $d_Y$ accordingly) for our method. In Table \ref{tab:dim_ablation} it can be seen that accuracy increases with the model size and that the number of FLOPs is low compared to the baseline, even for similar model sizes. As shown in Section \ref{method}, splitting the self-attention operator into two branches reduces the number of computations significantly, which in practice leads to a reduced number of FLOPs per forward-pass through the network. 

In Figure \ref{pareto}, we illustrate the trade-off between FLOPs and classification accuracy for different methods, highlighting the competitive trade-off our proposed method. This result is in agreement with recent trends in computer vision, where Transformers are shown to be more computationally efficient compared to other architectures \cite{han2021transformer}. \\~\

\noindent \textbf{Attention mechanisms} In order to demonstrate the effectiveness of the self-attention mechanisms, we ablate both the global and local attention modules in the Transformer blocks of the network by disabling the corresponding MSA operations. For the baseline method, there is only a single (global) attention mechanism which can be ablated. The results, shown in Table \ref{tab:att_ablation}, verify that for our proposed method, both modules are indeed useful and increase the overall accuracy. Furthermore, they are additive in the sense that they can be used interchangeably or in combination, in order to trade-off accuracy for computation and model size. The tighter bound on the accuracy when using global attention also suggests that it helps stabilize training. The baseline method also benefits from attention to some extent, but adding it to the network does not yield the same improvement as simply splitting the network into two branches. \\~\

\noindent \textbf{Number of anchors and neighbours} Finally, we investigate the effect of varying the number of anchor points and neighbours, as shown in Table \ref{tab:anchor_ablation}. As expected, a relatively large number of anchors is required in order to accurately represent the global shape of the point cloud. However, concerning the number of nearest neighbours, it seems that 10 is sufficient for good accuracy, but increasing it further reduces variance between different training runs.

\subsection{Feature Matching on 3DMatch}

In order to demonstrate how Point-TnT can be applied to a real-world scene resonstruction scenario, we consider the problem of feature matching on the 3DMatch dataset \cite{zeng20173dmatch}, which consists of 62 indoor scenes collected from RGB-D measurements. The goal of the feature matching task is to generate descriptors of local patches of the scenes which can be used for matching scans that have significant overlap. More specifically, given a pair of point cloud scenes $(\mathcal{X}, \mathcal{X'})$ with at least 30 \% overlap, we find the corresponding points by extracting local features and matching them using nearest neighbour search. It then becomes possible to register the two scenes by estimating a rigid transformation using e.g.\ RANSAC.

We use the official split, with 54 scenes for training and 8 for testing, and the same pre-processing setup as in DIP \cite{poiesi2021distinctive}, with the exception of the feature extraction network. Whereas the original implementation uses a Spatial Transformer for initial alignment of the point clouds and then a simple PointNet for feature extraction, we remove the Spatial Transformer entirely and replace the PointNet with our Point-TnT model. During training, we sample local patches consisting of $N = 256$ points from the overlapping regions, and train by minimizing the hardest contrastive loss \cite{choy2019fully}. We train our network for 10 epochs using the AdamW optimizer \cite{loshchilov2018decoupled} with a weight decay of 0.1 and a cosine learning rate schedule starting at 0.0001, without any data augmentation. Since each local patch processed by the network consists of only 256 points, we modify the parameters of our method to use $M = 48$ and $k=10$ anchor points and neighbours respectively. For fair comparison to DIP, we use the same feature dimension (32) at the final layer as the original implementation.

During evaluation, we randomly sample 5000 points from each fragment and report the feature matching recall (FMR), i.e.\ the fraction of successful alignments with an inlier ratio of at least 5 \%, where an inlier pair is defined by being less than 10 cm apart. In order to match corresponding patches, we use the global feature pairs $(Z, Z')$, where $Z$ and $Z'$ are calculated using (\ref{signature}) for the two patches, and perform mutual nearest neighbour search in feature space. We refer to the original DIP authors' publication for a complete description of the experimental settings and evaluation protocol \cite{poiesi2021distinctive}.

The results shown in Table \ref{tab:fmr} suggest that Point-TnT finds more descriptive features than the original DIP implementation and reduces the number of unsuccessful matches by 38 \%, without requiring any initial alignment, which shows that our method can be used for improving real-world scene reconstruction pipelines.

\begin{table}
	\caption{Feature matching recall and standard deviation on 3DMatch.}
	\centering
	\begin{tabular}{lcc}
	\toprule
	\textbf{Method} & FMR & std \\
	\midrule
	DIP \cite{poiesi2021distinctive} & 0.948 & 0.046 \\
	DIP + Point-TnT & 0.968 & 0.031 \\
	\bottomrule
	\end{tabular}
	\label{tab:fmr}
\end{table}

\section{Conclusions}

In this work, we have explored the limitations and advantages of the Transformer architecture for 3D shape recognition. We have shown that naively applying self-attention to all points in a point cloud is both computationally inefficient and not very useful for learning descriptive feature representations. However, when applied within and between local patches of points, representation ability is drastically improved. This result is in agreement with recent works in image classification \cite{dosovitskiy2021image, han2021transformer}, where the Transformer architecture has shown to work better on local image patches rather than individual pixels. It also makes feature extraction more computationally tractable, which creates new opportunities for applying attention mechanisms to edge use cases that rely on unstructured 3D data, such as simultaneous localization and mapping (SLAM), where computational resources are limited. In future work, we will consider integrating and extending our method for mobile SLAM pipelines. 

\section*{Acknowledgment}
This work was partially supported by the Wallenberg AI, Autonomous Systems and Software Program (WASP), funded by the Knut and Alice Wallenberg Foundation. 


\bibliographystyle{IEEEtran}
\bibliography{point_tnt_bib}

\begin{thebibliography}{10}
\providecommand{\url}[1]{#1}
\csname url@samestyle\endcsname
\providecommand{\newblock}{\relax}
\providecommand{\bibinfo}[2]{#2}
\providecommand{\BIBentrySTDinterwordspacing}{\spaceskip=0pt\relax}
\providecommand{\BIBentryALTinterwordstretchfactor}{4}
\providecommand{\BIBentryALTinterwordspacing}{\spaceskip=\fontdimen2\font plus
\BIBentryALTinterwordstretchfactor\fontdimen3\font minus
  \fontdimen4\font\relax}
\providecommand{\BIBforeignlanguage}[2]{{%
\expandafter\ifx\csname l@#1\endcsname\relax
\typeout{** WARNING: IEEEtran.bst: No hyphenation pattern has been}%
\typeout{** loaded for the language `#1'. Using the pattern for}%
\typeout{** the default language instead.}%
\else
\language=\csname l@#1\endcsname
\fi
#2}}
\providecommand{\BIBdecl}{\relax}
\BIBdecl

\bibitem{vaswani2017attention}
A.~Vaswani, N.~Shazeer, N.~Parmar, J.~Uszkoreit, L.~Jones, A.~N. Gomez,
  {\L}.~Kaiser, and I.~Polosukhin, ``{Attention is all you need},'' in
  \emph{Proceedings of the 31st International Conference on Neural Information
  Processing Systems}, 2017, pp. 6000--6010.

\bibitem{dosovitskiy2021image}
A.~Kolesnikov, A.~Dosovitskiy, D.~Weissenborn, G.~Heigold, J.~Uszkoreit,
  L.~Beyer, M.~Minderer, M.~Dehghani, N.~Houlsby, S.~Gelly, T.~Unterthiner, and
  X.~Zhai, ``An image is worth 16x16 words: Transformers for image recognition
  at scale,'' 2021.

\bibitem{touvron2021training}
H.~Touvron, M.~Cord, M.~Douze, F.~Massa, A.~Sablayrolles, and H.~J{\'e}gou,
  ``Training data-efficient image transformers \& distillation through
  attention,'' in \emph{International Conference on Machine Learning}.\hskip
  1em plus 0.5em minus 0.4em\relax PMLR, 2021, pp. 10\,347--10\,357.

\bibitem{carion2020end}
N.~Carion, F.~Massa, G.~Synnaeve, N.~Usunier, A.~Kirillov, and S.~Zagoruyko,
  ``End-to-end object detection with transformers,'' in \emph{European
  Conference on Computer Vision}.\hskip 1em plus 0.5em minus 0.4em\relax
  Springer, 2020, pp. 213--229.

\bibitem{neimark2021video}
D.~Neimark, O.~Bar, M.~Zohar, and D.~Asselmann, ``Video transformer network,''
  in \emph{Proceedings of the IEEE/CVF International Conference on Computer
  Vision}, 2021, pp. 3163--3172.

\bibitem{gulati2020conformer}
A.~Gulati, J.~Qin, C.-C. Chiu, N.~Parmar, Y.~Zhang, J.~Yu, W.~Han, S.~Wang,
  Z.~Zhang, Y.~Wu \emph{et~al.}, ``{Conformer: Convolution-augmented
  Transformer for Speech Recognition},'' \emph{Proc. Interspeech 2020}, pp.
  5036--5040, 2020.

\bibitem{chen2021developing}
X.~Chen, Y.~Wu, Z.~Wang, S.~Liu, and J.~Li, ``Developing real-time streaming
  transformer transducer for speech recognition on large-scale dataset,'' in
  \emph{ICASSP 2021-2021 IEEE International Conference on Acoustics, Speech and
  Signal Processing (ICASSP)}.\hskip 1em plus 0.5em minus 0.4em\relax IEEE,
  2021, pp. 5904--5908.

\bibitem{liu2021tera}
A.~T. Liu, S.-W. Li, and H.-y. Lee, ``Tera: Self-supervised learning of
  transformer encoder representation for speech,'' \emph{IEEE/ACM Transactions
  on Audio, Speech, and Language Processing}, vol.~29, pp. 2351--2366, 2021.

\bibitem{berg21_interspeech}
A.~Berg, M.~O’Connor, and M.~T. Cruz, ``{Keyword Transformer: A
  Self-Attention Model for Keyword Spotting},'' in \emph{Proc. Interspeech
  2021}, 2021, pp. 4249--4253.

\bibitem{lee2019set}
J.~Lee, Y.~Lee, J.~Kim, A.~Kosiorek, S.~Choi, and Y.~W. Teh, ``Set transformer:
  A framework for attention-based permutation-invariant neural networks,'' in
  \emph{International Conference on Machine Learning}.\hskip 1em plus 0.5em
  minus 0.4em\relax PMLR, 2019, pp. 3744--3753.

\bibitem{zhao2021point}
H.~Zhao, L.~Jiang, J.~Jia, P.~H. Torr, and V.~Koltun, ``Point transformer,'' in
  \emph{Proceedings of the IEEE/CVF International Conference on Computer
  Vision}, 2021, pp. 16\,259--16\,268.

\bibitem{guo2021pct}
M.-H. Guo, J.-X. Cai, Z.-N. Liu, T.-J. Mu, R.~R. Martin, and S.-M. Hu, ``Pct:
  Point cloud transformer,'' \emph{Computational Visual Media}, vol.~7, no.~2,
  pp. 187--199, 2021.

\bibitem{han2021transformer}
K.~Han, A.~Xiao, E.~Wu, J.~Guo, C.~Xu, and Y.~Wang, ``Transformer in
  transformer,'' \emph{Advances in Neural Information Processing Systems},
  vol.~34, 2021.

\bibitem{uy2019revisiting}
M.~A. Uy, Q.-H. Pham, B.-S. Hua, T.~Nguyen, and S.-K. Yeung, ``Revisiting point
  cloud classification: A new benchmark dataset and classification model on
  real-world data,'' in \emph{Proceedings of the IEEE/CVF International
  Conference on Computer Vision}, 2019, pp. 1588--1597.

\bibitem{wu20153d}
Z.~Wu, S.~Song, A.~Khosla, F.~Yu, L.~Zhang, X.~Tang, and J.~Xiao, ``3d
  shapenets: A deep representation for volumetric shapes,'' in
  \emph{Proceedings of the IEEE conference on computer vision and pattern
  recognition}, 2015, pp. 1912--1920.

\bibitem{maturana2015voxnet}
D.~Maturana and S.~Scherer, ``Voxnet: A 3d convolutional neural network for
  real-time object recognition,'' in \emph{2015 IEEE/RSJ International
  Conference on Intelligent Robots and Systems (IROS)}.\hskip 1em plus 0.5em
  minus 0.4em\relax IEEE, 2015, pp. 922--928.

\bibitem{choy20194d}
C.~Choy, J.~Gwak, and S.~Savarese, ``4d spatio-temporal convnets: Minkowski
  convolutional neural networks,'' in \emph{Proceedings of the IEEE/CVF
  Conference on Computer Vision and Pattern Recognition}, 2019, pp. 3075--3084.

\bibitem{su2015multi}
H.~Su, S.~Maji, E.~Kalogerakis, and E.~Learned-Miller, ``Multi-view
  convolutional neural networks for 3d shape recognition,'' in
  \emph{Proceedings of the IEEE international conference on computer vision},
  2015, pp. 945--953.

\bibitem{hamdi2021mvtn}
A.~Hamdi, S.~Giancola, and B.~Ghanem, ``Mvtn: Multi-view transformation network
  for 3d shape recognition,'' in \emph{Proceedings of the IEEE/CVF
  International Conference on Computer Vision}, 2021, pp. 1--11.

\bibitem{wei2020view}
X.~Wei, R.~Yu, and J.~Sun, ``View-gcn: View-based graph convolutional network
  for 3d shape analysis,'' in \emph{Proceedings of the IEEE/CVF Conference on
  Computer Vision and Pattern Recognition}, 2020, pp. 1850--1859.

\bibitem{zaheer2017deep}
M.~Zaheer, S.~Kottur, S.~Ravanbhakhsh, B.~P{\'o}czos, R.~Salakhutdinov, and
  A.~J. Smola, ``Deep sets,'' in \emph{Proceedings of the 31st International
  Conference on Neural Information Processing Systems}, 2017, pp. 3394--3404.

\bibitem{qi2017pointnet}
C.~R. Qi, H.~Su, K.~Mo, and L.~J. Guibas, ``Pointnet: Deep learning on point
  sets for 3d classification and segmentation,'' in \emph{Proceedings of the
  IEEE conference on computer vision and pattern recognition}, 2017, pp.
  652--660.

\bibitem{jaderberg2015spatial}
M.~Jaderberg, K.~Simonyan, A.~Zisserman \emph{et~al.}, ``Spatial transformer
  networks,'' \emph{Advances in neural information processing systems},
  vol.~28, pp. 2017--2025, 2015.

\bibitem{qi2017pointnet++}
C.~R. Qi, L.~Yi, H.~Su, and L.~J. Guibas, ``Pointnet++: Deep hierarchical
  feature learning on point sets in a metric space,'' \emph{Advances in Neural
  Information Processing Systems}, vol.~30, 2017.

\bibitem{wang2019dynamic}
Y.~Wang, Y.~Sun, Z.~Liu, S.~E. Sarma, M.~M. Bronstein, and J.~M. Solomon,
  ``Dynamic graph cnn for learning on point clouds,'' \emph{Acm Transactions On
  Graphics (tog)}, vol.~38, no.~5, pp. 1--12, 2019.

\bibitem{chen2020dapnet}
L.~Chen, Z.~Xu, Y.~Fu, H.~Huang, S.~Wang, and H.~Li, ``Dapnet: A double
  self-attention convolutional network for segmentation of point clouds,''
  \emph{arXiv preprint arXiv:2004.08596}, 2020.

\bibitem{hendrycks2016gaussian}
D.~Hendrycks and K.~Gimpel, ``{Gaussian error linear units (GELUs)},''
  \emph{arXiv preprint arXiv:1606.08415}, 2016.

\bibitem{ba2016layer}
J.~L. Ba, J.~R. Kiros, and G.~E. Hinton, ``{Layer normalization},'' \emph{arXiv
  preprint arXiv:1607.06450}, 2016.

\bibitem{he2016deep}
K.~He, X.~Zhang, S.~Ren, and J.~Sun, ``{Deep residual learning for image
  recognition},'' in \emph{Proceedings of the IEEE conference on computer
  vision and pattern recognition}, 2016, pp. 770--778.

\bibitem{eldar1997farthest}
Y.~Eldar, M.~Lindenbaum, M.~Porat, and Y.~Y. Zeevi, ``The farthest point
  strategy for progressive image sampling,'' \emph{IEEE Transactions on Image
  Processing}, vol.~6, no.~9, pp. 1305--1315, 1997.

\bibitem{xu2018spidercnn}
Y.~Xu, T.~Fan, M.~Xu, L.~Zeng, and Y.~Qiao, ``Spidercnn: Deep learning on point
  sets with parameterized convolutional filters,'' in \emph{Proceedings of the
  European Conference on Computer Vision (ECCV)}, 2018, pp. 87--102.

\bibitem{li2018pointcnn}
Y.~Li, R.~Bu, M.~Sun, W.~Wu, X.~Di, and B.~Chen, ``Pointcnn: Convolution on
  x-transformed points,'' \emph{Advances in neural information processing
  systems}, vol.~31, pp. 820--830, 2018.

\bibitem{mazur2021cloud}
K.~Mazur and V.~Lempitsky, ``Cloud transformers: A universal approach to point
  cloud processing tasks,'' in \emph{Proceedings of the IEEE/CVF International
  Conference on Computer Vision}, 2021, pp. 10\,715--10\,724.

\bibitem{goyal2021revisiting}
A.~Goyal, H.~Law, B.~Liu, A.~Newell, and J.~Deng, ``Revisiting point cloud
  shape classification with a simple and effective baseline,''
  \emph{International Conference on Machine Learning}, 2021.

\bibitem{qiu2021geometric}
S.~Qiu, S.~Anwar, and N.~Barnes, ``Geometric back-projection network for point
  cloud classification,'' \emph{IEEE Transactions on Multimedia}, 2021.

\bibitem{loshchilov2018decoupled}
I.~Loshchilov and F.~Hutter, ``Decoupled weight decay regularization,'' in
  \emph{International Conference on Learning Representations}, 2018.

\bibitem{lee2021regularization}
D.~Lee, J.~Lee, J.~Lee, H.~Lee, M.~Lee, S.~Woo, and S.~Lee, ``Regularization
  strategy for point cloud via rigidly mixed sample,'' in \emph{Proceedings of
  the IEEE/CVF Conference on Computer Vision and Pattern Recognition}, 2021,
  pp. 15\,900--15\,909.

\bibitem{zeng20173dmatch}
A.~Zeng, S.~Song, M.~Nie{\ss}ner, M.~Fisher, J.~Xiao, and T.~Funkhouser,
  ``3dmatch: Learning local geometric descriptors from rgb-d reconstructions,''
  in \emph{Proceedings of the IEEE conference on computer vision and pattern
  recognition}, 2017, pp. 1802--1811.

\bibitem{poiesi2021distinctive}
F.~Poiesi and D.~Boscaini, ``Distinctive 3d local deep descriptors,'' in
  \emph{2020 25th International Conference on Pattern Recognition
  (ICPR)}.\hskip 1em plus 0.5em minus 0.4em\relax IEEE, 2021, pp. 5720--5727.

\bibitem{choy2019fully}
C.~Choy, J.~Park, and V.~Koltun, ``Fully convolutional geometric features,'' in
  \emph{Proceedings of the IEEE/CVF International Conference on Computer
  Vision}, 2019, pp. 8958--8966.

\end{thebibliography}

\end{document}